\documentclass{article}

\UseRawInputEncoding

\usepackage{PRIMEarxiv}

\usepackage[T1]{fontenc}
\usepackage{helvet}      

\usepackage[table,dvipsnames]{xcolor}

\usepackage{hyperref}
\usepackage{url}
\usepackage{booktabs}
\usepackage{amsfonts}
\usepackage{amssymb}
\usepackage{nicefrac}
\usepackage{microtype}
\usepackage{graphicx}
\usepackage{multirow}
\usepackage{array}
\usepackage{siunitx}
\usepackage{listings}
\usepackage{enumitem}
\usepackage{tabularx}
\newcolumntype{C}{>{\centering\arraybackslash}X}
\tcbuselibrary{listings}

\lstdefinelanguage{yaml}{
  morekeywords={true,false,null,y,n},
  sensitive=false,
  morecomment=[l]{\#},
  morestring=[b]",
  morestring=[b]',
  alsoletter={-},
  moredelim=[l][\color{black}\ttfamily]{-},
}

\definecolor{lstbg}{HTML}{F8FAFC}
\definecolor{lstkeyword}{HTML}{1E40AF}
\definecolor{lststring}{HTML}{0D9488}
\definecolor{lstcomment}{HTML}{64748B}
\definecolor{lstfunc}{HTML}{9333EA}
\definecolor{lsttitlebg}{HTML}{1E293B}
\definecolor{linkblue}{HTML}{1E40AF}
\definecolor{urlteal}{HTML}{0D9488}

\hypersetup{
    pdftitle={CuratorKIT : Data Curation and Synthetic Data Generation for LLM Post-Training},
    pdfauthor={Soham Bhattacharjee, Karun Sharma, Vinay Kumar Sankarapu, Pratinav Seth},
    pdfsubject={Synthetic Data Curation for LLM Post-Training},
    pdfkeywords={Synthetic Data, Data Curation, LLM Post-Training, LLM-as-Judge,
                 Hallucination Detection, Data Quality, Open-Source},
    colorlinks=true,
    linkcolor=linkblue,
    citecolor=linkblue,
    urlcolor=urlteal,
}

\newcommand{\cmark}{{\color{green!45!black}\checkmark}}
\newcommand{\xmark}{{\color{black!25}\texttimes}}
\newcommand{\tmark}{{\color{orange!55!black}\textasciitilde}}

\setlist[itemize]{parsep=0pt, itemsep=3pt}
\setlist[enumerate]{parsep=0pt, itemsep=3pt}

\newtcblisting{codeblock}[1][]{
    enhanced,
    listing only,
    colback=lstbg,
    colframe=lsttitlebg,
    boxrule=0.25pt,
    arc=3pt,
    outer arc=3pt,
    top=4pt, bottom=2pt, left=0pt, right=0pt,
    before skip=4pt,
    after skip=4pt,
    title=\hspace{2pt},
    fonttitle=\ttfamily\tiny\color{lstcomment},
    coltitle=lstcomment,
    colbacktitle=lsttitlebg,
    attach boxed title to top left={yshift=0pt, xshift=0pt},
    boxed title style={
        size=minimal,
        height=5pt,
        width=\textwidth+0.5pt,
        arc=3pt,
        bottom=0pt,
        colback=lsttitlebg,
        boxrule=0pt,
    },
    listing options={
        basicstyle=\footnotesize\ttfamily\color{black!80},
        keywordstyle=\color{lstkeyword}\bfseries,
        commentstyle=\color{lstcomment}\itshape,
        stringstyle=\color{lststring},
        identifierstyle=\color{black},
        emph={CuratorConfig,Curator,CuratorResult,DataSample,RejectedSample,HallucinationGate,RewardGate,DiversityGate,SchemaGate,Pipeline,run,apply_patch,diagnose,generate,BaseGenerationTask,BaseGate,BaseLLM,LLMResponse,ProvenanceRecord},
        emphstyle=\color{lstfunc}\bfseries,
        numbers=none,
        breaklines=true,
        breakatwhitespace=true,
        showstringspaces=false,
        tabsize=4,
        xleftmargin=0.3em,
        columns=flexible,
        keepspaces=true,
        lineskip=0.5pt,
        aboveskip=2pt,
        belowskip=2pt,
    },
    #1,
}

\title{\includegraphics[width=1\linewidth]{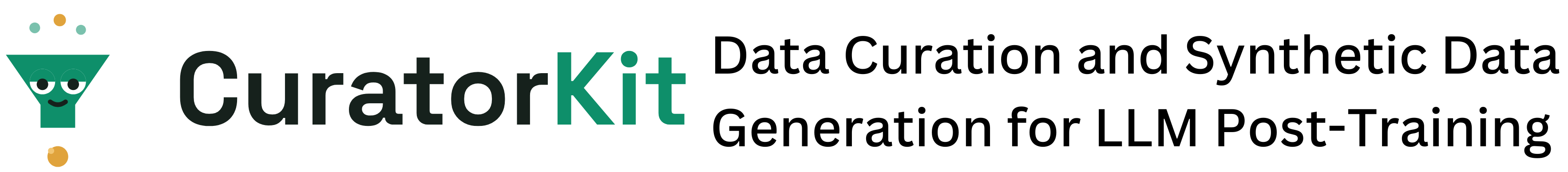}}

\author{
  Soham Bhattacharjee, Karun Sharma, \\
  Vinay Kumar Sankarapu, Pratinav Seth \\
  Lexsi Labs \\
  \texttt{pratinav.seth@lexsi.ai}
}

\setabstract{%
Data curation is a critical part of post-training pipelines for large language models, yet existing tools often treat ingestion, deduplication, synthetic generation, and quality filtering as separate stages. This fragmentation makes it difficult to audit pipeline decisions or understand why individual samples are rejected. CuratorKIT is an open-source Python library that covers this full lifecycle in a single configurable pipeline. The framework is composed of six source format readers and automatic schema detection, a pre-generation data hygiene layer for credentials, PII, and toxic content, eight LLM-powered generation tasks, three complementary quality gates with provenance-exact hallucination verification, structured adaptive recovery, and five training-ready export formats compatible with TRL, Unsloth, and AlignTune. Every pipeline decision is recorded in an append-only per-sample provenance chain, and rejected samples carry structured failure reasons rather than being silently discarded. CuratorKIT supports 100+ LLM providers through LiteLLM, exposes both a Python API and a YAML-driven CLI, and is designed for practitioners who need reproducible, auditable data pipelines at scale .
}

\setkeywords{Synthetic Data, Data Curation, LLM Post-Training, LLM-as-Judge,
  Hallucination Detection, Data Quality, Open-Source Framework}

\runningtitle{CuratorKIT : Data Curation and Synthetic Data Generation for LLM Post-Training}

\date{June 2026}

\begin{document}
\maketitle

\section{Introduction}
\label{sec:introduction}

Building high-quality post-training datasets for large language models
requires ingestion from heterogenous sources, removing duplicates, generating or transforming samples, applying quality filters and exporting the resultant dataset in a format suitable for training. In
practice, most projects assemble this chain from independent scripts with
incompatible schemas, no shared audit trail, and no mechanism for understanding
why a sample was rejected or whether it could have been recovered. This makes the resulting pipeline difficult to
reproduce, difficult to inspect, and poorly suited for iterative improvement.

Existing frameworks address parts of this problem. distilabel~\cite{distilabel}
and AgentInstruct~\cite{agentinstruct} cover LLM-powered generation and
preference scoring. DataTrove~\cite{penedo2024datatrove} handles large-scale
ingestion and cleaning. NeMo Data Designer~\cite{nemo-data-designer} targets
structured synthetic generation from templates. While these tools are helpful, they do not offer a unified lifecycle that connects source ingestion, generation, validation, recovery, and export under a single audit model. The absence of provenance is particularly consequential for hallucination gating. If the source evidence associated with a sample
is not preserved, a later verifier must retrieve an approximately relevant
chunk post-hoc during quality filtering. That retrieval step can introduce errors precisely where
the pipeline requires the strongest grounding. Similarly, when rejected samples
are treated as discarded noise rather than recoverable artifacts, pipelines lose
the opportunity to diagnose systematic failures and apply targeted repair.

We introduce CuratorKIT, a unified and configurable library for building
auditable post-training data curation pipelines. CuratorKIT covers the full curation lifecycle in a single framework, spanning synthetic data generation, quality filtering, schema validation, and training-ready export. Every processing
decision is recorded in an append-only per-sample provenance chain, and rejected
samples retain structured failure reasons instead of being silently removed. It supports 100+
LLM providers through LiteLLM~\cite{litellm2024}, provides both a Python API and
a YAML-driven CLI, and is designed for practitioners who need reproducible,
auditable data pipelines at scale. The empirical validation of its core
gating and recovery components is reported in the companion
study~\cite{curatorkit_emnlp26}.
This document describes the full library: its architecture, every component,
and how to use it.

\section{Architecture}
\label{sec:architecture}

\begin{figure}[t]
  \centering
  \includegraphics[width=0.99\textwidth]{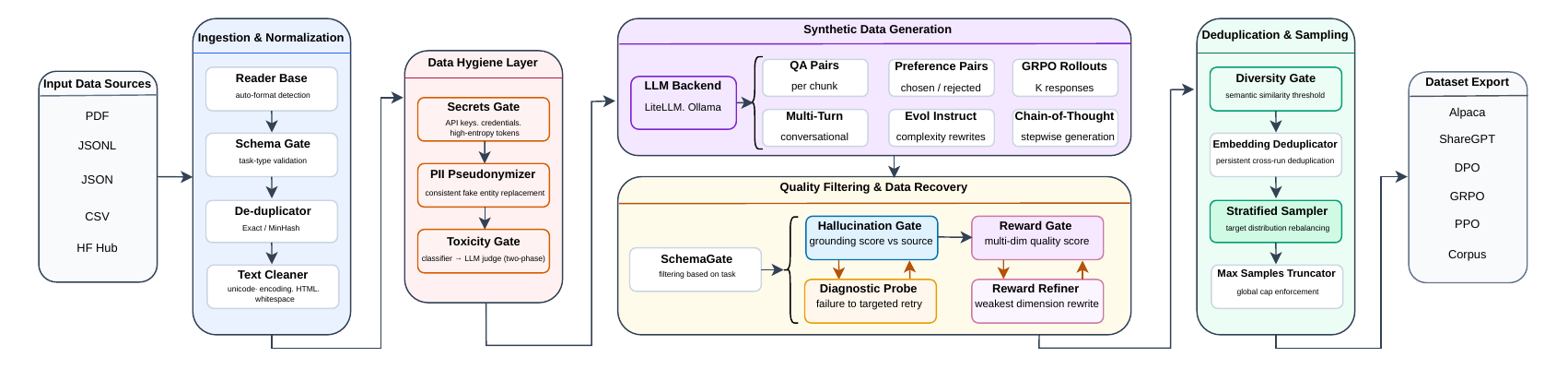}
  \caption{CuratorKIT pipeline. Data flows from six source formats through
    ingestion, optional generation, quality gating, and adaptive recovery.
    Accepted samples are exported in five training formats.}
  \label{fig:pipeline}
\end{figure}

CuratorKIT follows a pipeline architecture where data flows through a
sequence of steps, as shown in Figure~\ref{fig:pipeline}. Each step is one
of four types (reader, gate, normalizer, or exporter), dispatched by type in
a runner that is approximately 100 lines of code. Adding a new step means
implementing one of four abstract base classes.

\subsection{The DataSample}

Every record in the pipeline is a \texttt{DataSample}: a Pydantic model with
thirteen fields covering all eight supported task types. Rather than
maintaining separate schemas per task family, CuratorKIT uses a single
unified record with fields activated or ignored based on the
\texttt{task\_type} tag.

\begin{table}[ht]
  \centering
  \small
  \caption{\texttt{DataSample} fields. The \texttt{task\_type} tag governs
    which fields are required and how they are validated.}
    \vspace{10pt}
  \label{tab:schema}
  \begin{tabularx}{\textwidth}{lX}
    \toprule
    \textbf{Field} & \textbf{Purpose} \\
    \midrule
    \texttt{id} & Stable UUID4 identifier \\
    \texttt{source\_uri} & Source document or file URI \\
    \texttt{instruction} & Prompt or instruction \\
    \texttt{input} & Context passage (Alpaca-style); also stores source
      chunk for grounded generation \\
    \texttt{output} & Target response (SFT); full text (pretraining) \\
    \texttt{chosen} / \texttt{rejected} & DPO preference pair completions \\
    \texttt{label} & Scalar quality score (unpaired preference) \\
    \texttt{responses} / \texttt{reward\_scores} & GRPO rollouts and scores \\
    \texttt{task\_type} & One of eight task family tags \\
    \texttt{metadata} & Arbitrary step-specific data \\
    \texttt{provenance\_chain} & Append-only list of step records \\
    \bottomrule
  \end{tabularx}
\end{table}

\subsection{Provenance and Rejection}

Each sample carries an append-only \texttt{provenance\_chain}, which stores
the sequence of \texttt{ProvenanceRecord} objects added by pipeline steps.
Each record contains the step name, a version string, a UTC timestamp, a
configuration hash, and a free-form \texttt{notes} dictionary. Once a record
is appended, it is never modified. This design makes pipeline decisions
auditable at the level of individual samples. For example, a sample that fails
the hallucination gate retains both the grounding score and the judge's
specific complaints in its provenance chain.

A \texttt{RejectedSample} extends \texttt{DataSample} with two additional
fields: \texttt{rejection\_reason}, a structured string such as
\texttt{"hallucination\_contract\_failed:0.42"}, and
\texttt{rejecting\_step}, the class that rejected the sample. Any step that
drops a sample must return a \texttt{RejectedSample} with a structured reason;
silent drops are treated as bugs. After each run, all rejected samples are
written to \texttt{rejected.jsonl}.

\subsection{Configuration and API}

All pipeline behaviour is driven by a single \texttt{CuratorConfig}
dataclass, a flat configuration object with over 100 fields spanning data
sources, LLM backends, generation tasks, quality gates, and hygiene settings.
There are no nested sub-objects, no hidden state, and no step-wiring code to
write. The entry point mirrors the TRL~\cite{trl2024} trainer pattern, with
\texttt{CuratorConfig} serving the role that \texttt{TrainingArguments}
plays in HuggingFace Transformers~\cite{transformers2024}:

\begin{codeblock}[]
from curatorkit import Curator, CuratorConfig

config = CuratorConfig(
    dataset="data/corpus.jsonl",
    generation_task="qa",
    llm_model="openai/gpt-4o-mini",
    hallucination_threshold=0.7,
    reward_threshold=0.7,
    export_formats=["alpaca", "sharegpt"],
)

curator = Curator(config)
result = curator.run()
# result.passed     — accepted DataSample objects
# result.rejected   — RejectedSample objects with structured reasons
# result.stage_counts — per-step input/output/rejected tallies
# result.diagnostics  — PipelineDiagnostics (when probe is active)
\end{codeblock}

The \texttt{apply\_patch(dict)} method returns a shallow copy of the config
with only the specified fields changed, leaving the original unchanged. This
powers the adaptive recovery system, which tests repair strategies without
mutating the base configuration.

\subsection{Mandatory Outputs}

Every pipeline run produces four outputs:

\begin{itemize}
  \item \texttt{manifest.json}: configuration hash, per-stage counts,
    rejection breakdown, dedup statistics, token usage, wall-clock time.
  \item \texttt{dataset\_card.md}: human-readable Markdown summary with
    stage-by-stage pass rates and reproduction instructions.
  \item \texttt{rejected.jsonl}: one JSON object per rejected sample with
    the full provenance chain and structured rejection reason.
  \item \texttt{checksums.txt}: SHA-256 hashes of all output files.
\end{itemize}

These mandatory outputs since they track the reproducibility and summary of the generation. They are written after every run regardless of which
steps were configured.

\section{Ingestion \& Normalisation}
\label{sec:ingestion}

The ingestion and normalization layer handles the non-generative stages of the
curation pipeline. It reads raw data from heterogeneous source formats, resolves
schema ambiguities, enforces structural constraints, removes redundant samples,
cleans text, and rebalances distributions before downstream use. No LLM call is
made at any point in this layer. The design follows a single invariant: every
record that enters the pipeline must exit either as a well-formed
\texttt{DataSample} or as a \texttt{RejectedSample} with a structured reason.

\subsection{Source Format Readers}
\label{sec:source-format-readers}

CuratorKIT provides six source format readers, each implemented as a subclass
of \texttt{BaseConnector}. A reader subclass is responsible for implementing a
single method, \texttt{\_iter\_rows()}, which yields
\texttt{(line\_no, raw\_dict)} tuples. The base connector then handles the
shared downstream logic, including user preprocessing, field mapping, format
detection, \texttt{DataSample} construction, and rejection on parse failure.
This separation keeps source-specific parsing separate from the rest of the
pipeline, so adding a new source format does not require changes to downstream
components.

CuratorKIT includes six readers for common file, dataset, and document formats.
It supports line-delimited JSON, JSON arrays with an optional
\texttt{data\_key}, CSV and TSV files, Apache Parquet files, Hugging Face Hub
datasets, and PDFs. The CSV reader includes header detection and JSON cell
parsing, the Hugging Face reader supports streaming, subsets, and splits, and
the PDF reader uses MinerU for layout-preserving extraction with multiple
chunking strategies and optional table extraction.

\textbf{Field mapping.}
Source column names rarely match the canonical \texttt{DataSample} field
names. The \texttt{field\_mapping} parameter handles this translation, with no preprocessing code needed for column renaming:

\begin{codeblock}[]
CuratorConfig(
    dataset="data/raw.jsonl",
    field_mapping={
        "instruction": "question",    # source column -> DataSample field
        "output":      "answer",
        "input":       "context",
    },
)
\end{codeblock}

\medskip

\textbf{User preprocessing.}
For transformations beyond column renaming, a \texttt{preprocessing\_fn}
callable runs on every raw row before field mapping and format detection.
Returning \texttt{None} drops the row; returning a fully-constructed
\texttt{DataSample} bypasses all subsequent detection logic, giving the
caller complete control when the automatic path is insufficient:

\begin{codeblock}[]
def clean_row(row: dict) -> dict | None:
    if len(row.get("answer", "")) < 20:
        return None                        # drop short answers
    row["answer"] = " ".join(row["answer"].split())
    return row

CuratorConfig(
    dataset="data/raw.jsonl",
    preprocessing_fn=clean_row,
    field_mapping={"instruction": "question", "output": "answer"},
)
\end{codeblock}

\medskip

\textbf{Multi-source pipelines.}
The \texttt{dataset} parameter accepts a list of sources, where each
entry can be a string path, a HuggingFace dataset identifier, or a dict
with per-source overrides for split, subset, and sample cap. A
top-level \texttt{split} applies as the default for any source that does
not specify its own:

\begin{codeblock}[]
CuratorConfig(
    dataset=[
        "tatsu-lab/alpaca",
        "data/extra.jsonl",
        {"name": "openai/summarize_from_feedback",
         "split": "validation", "max_samples": 500},
    ],
    split="train",
)
\end{codeblock}

\medskip

\textbf{PDF chunking.}
The PDF reader exposes the full MinerU extraction surface through
\texttt{CuratorConfig}: chunking strategy, token cap, overlap, and table
extraction are all configurable. The \texttt{pdf\_output\_mode} parameter
allows in-reader generation (\texttt{"qa"}, \texttt{"preference"},
\texttt{"grpo"}, \texttt{"multiturn"}), enabling a pipeline that starts
from a raw PDF corpus and produces training-ready samples without any
intermediate file:

\begin{codeblock}[]
CuratorConfig(
    dataset="docs/contracts.pdf",
    pdf_chunk_strategy="heading",
    pdf_chunk_max_tokens=512,
    pdf_chunk_overlap_tokens=50,
    pdf_extract_tables=True,
    pdf_output_mode="qa",    # or "chunk" to defer generation
)
\end{codeblock}

\subsection{Format Auto-Detection}

Real-world datasets rarely follow a single column schema. The same field may
appear under different names across datasets, and some column names are
ambiguous across task types. To reduce manual configuration, CuratorKIT uses a
three-stage detection procedure to infer the format of each source.

The first stage builds candidate mappings from the observed columns. CuratorKIT
maps more than 60 common column-name variants to semantic slots. For example,
\texttt{instruction}, \texttt{prompt}, \texttt{query}, \texttt{question},
\texttt{user\_input}, and \texttt{human} can all map to the instruction field.
When a name is ambiguous, such as \texttt{input}, CuratorKIT applies a
deterministic priority order so that detection remains reproducible.

The second stage validates each candidate against the actual values in the
data. For example, a \texttt{conversations} column is accepted only if it has
the expected conversational structure, such as \texttt{list[dict]}. If the
column instead contains a flat string, that candidate is rejected and the
next-best mapping is evaluated.

The third stage normalizes role names for chat-style datasets. Aliases such as
\texttt{human}, \texttt{user}, and \texttt{input} are mapped to
\texttt{"user"}, while \texttt{assistant}, \texttt{gpt}, and \texttt{model}
are mapped to \texttt{"assistant"}. This supports both ChatML and ShareGPT-style
conventions without requiring users to rewrite their input data.

After inspecting the first $N$ rows, where $N$ is configurable and defaults to
10, the detector selects a format and reports a confidence tier among
\texttt{HIGH}, \texttt{MEDIUM}, \texttt{LOW}, or \texttt{UNKNOWN}. This makes
automatic detection transparent because users can see when the inferred schema is
reliable and when it may need manual review.

\subsection{Schema Validation}

The \texttt{SchemaGate} is the first gate in the pipeline and the only
one that is always active. It validates field presence, token count
bounds, and encoding for every sample, with field requirements
determined by the sample's \texttt{task\_type} tag: SFT samples
(\texttt{instruction\_following}, \texttt{conversational}) require
\texttt{instruction} and \texttt{output}; preference samples require
\texttt{chosen} and \texttt{rejected}; GRPO samples require
\texttt{instruction} and a non-empty \texttt{responses} list; language
modeling samples require only \texttt{output}. Token counting is
correspondingly task-aware, counting \texttt{instruction + chosen} for
preference pairs and \texttt{instruction + longest\_response} for GRPO
rollouts. Both whitespace-based counting (default) and
\texttt{tiktoken} tokenisation are supported.

\begin{codeblock}[]
CuratorConfig(
    min_tokens=10,
    max_tokens=4096,
    use_tiktoken=False,    # True for tiktoken cl100k_base
    schema_gate=True,      # False to skip all schema checks
)
\end{codeblock}

\subsection{Text Cleaning}

\texttt{TextCleaner} provides five independently selectable transforms:
HTML tag stripping, Unicode normalisation (NFC), mojibake encoding
repair, whitespace collapse, and control character removal. Each
transform appends a record to the sample's provenance chain, so the
exact sequence of cleaning operations applied to any sample is
recoverable from the audit log itself.

\begin{codeblock}[]
CuratorConfig(
    clean=True,
    clean_transforms=["html", "unicode", "whitespace"],
)
\end{codeblock}

\subsection{Deduplication}

Three complementary deduplication strategies address redundancy at
different scales and across different notions of similarity. All three
are task-type-aware: the hash key or embedding input is constructed from
the semantically meaningful fields for the given task family, rather
than a fixed set of columns.

\textbf{ExactDeduplicator} uses SHA-256 hashing of normalised text to
remove identical samples before they reach the generation or export
stages. The hash key is constructed from the fields that carry the
sample's full semantic content, which varies by task type. SFT,
evol-instruct, and chain-of-thought samples are hashed over the
instruction and output together, preference pairs additionally include
the chosen and rejected completions, and GRPO rollouts are hashed over
the instruction alongside all responses in the rollout.

\textbf{MinHashDeduplicator} removes near-duplicates using character
$n$-gram MinHash, with $n=3$ by default. It uses LSH banding to avoid comparing
every sample against every other sample, making near-duplicate detection more
efficient on larger datasets. The banding parameters are chosen automatically
from the configured similarity threshold, which defaults to 0.85.

\textbf{EmbeddingDeduplicator} supports semantic and cross-run deduplication by
maintaining a persistent embedding index on disk. When FAISS~\cite{douze2025faisslibrary}
is available, it is used for approximate nearest-neighbor search; otherwise,
CuratorKIT falls back to a NumPy-based brute-force search. Because the index is
saved across runs, samples accepted in earlier runs are not reintroduced when
the pipeline is run again on newly added data.

\begin{codeblock}[]
CuratorConfig(
    dedup="minhash",
    minhash_threshold=0.85,
    embedding_dedup=True,
    embedding_index_dir="output/embedding_index",
    embedding_dedup_threshold=0.92,
)
\end{codeblock}

\subsection{Sampling, Truncation, and Export}

\textbf{StratifiedSampler} corrects skewed category distributions by
downsampling over-represented categories to their configured target
fractions. Under-represented categories are never silently accepted:
they produce an explicit manifest warning that names the affected
categories and the magnitude of the shortfall, making distribution
imbalances visible in the audit record rather than absorbed quietly.

\textbf{MaxSamplesTruncator} enforces a global output cap after all
resampling has completed. Positioning truncation after
\texttt{StratifiedSampler} ensures that the target distribution is
established before the size limit is applied, rather than cutting into
a distribution that has not yet been rebalanced. When a per-source cap
is needed before samples from different readers are pooled, the
per-reader \texttt{max\_samples} override is the correct mechanism.

\begin{codeblock}[]
CuratorConfig(
    resample=True,
    resample_field="source_dataset",
    target_distribution={"legal": 0.4, "medical": 0.3, "general": 0.3},
    max_samples=10000,
)
\end{codeblock}
\vspace{10mm}
CuratorKIT exports to five training-ready formats:
\texttt{AlpacaExporter} ($\rightarrow$ \texttt{sft\_alpaca.jsonl}),
\texttt{ShareGPTExporter} ($\rightarrow$ \texttt{sft\_sharegpt.jsonl},
with full multi-turn support), \texttt{DPOExporter}
($\rightarrow$ \texttt{dpo.jsonl}, auto-detecting conversational turns),
\texttt{GRPOExporter} ($\rightarrow$ \texttt{grpo.jsonl}),
\texttt{PPOExporter} ($\rightarrow$ \texttt{ppo.jsonl}), and a
\texttt{CorpusExporter} ($\rightarrow$ \texttt{corpus.jsonl}) that
preserves full chunk metadata for downstream corpus analysis. Multiple
formats can be specified in a single run. Output splitting via
\texttt{output\_split} shuffles accepted samples and writes each
format's output into per-split subdirectories:

\begin{codeblock}[]
CuratorConfig(
    export_formats=["alpaca", "sharegpt", "dpo"],
    output_split={"train": 0.8, "val": 0.1, "test": 0.1},
    output_dir="output/",
)
\end{codeblock}

\section{Data Hygiene Layer}
\label{sec:hygiene}

The data hygiene layer comprises three components that clean source data
before any LLM generation call is made: \texttt{SecretsGate},
\texttt{PIIPseudonymizer}, and \texttt{ToxicityGate}. Their placement before
generation is a deliberate constraint: credentials are rejected before a
generation request is issued, PII is pseudonymised before a model generates
continuations, and toxic material is discarded before paying for inference.
All three are task-type-aware, selecting the fields appropriate for each
sample's \texttt{task\_type} rather than scanning a fixed column set, with
the option to override field selection explicitly. Execution order is fixed,
\texttt{SecretsGate} $\rightarrow$ \texttt{PIIPseudonymizer} $\rightarrow$
\texttt{ToxicityGate}, ordered by increasing computational cost so that the
cheapest gate clears the most obvious contamination before the more expensive
components run.

\subsection{SecretsGate}

\texttt{SecretsGate} rejects samples containing credentials, API keys, or
high-entropy tokens using the \texttt{detect-secrets} battery: deterministic
regex plugins covering AWS keys, GitHub tokens, PEM private keys, JWTs, and
common SaaS tokens, combined with Shannon entropy analysis for Base64 and
hexadecimal strings. The design rejects rather than redacts, as retaining
the surrounding context of a redacted credential still teaches a fine-tuned
model the structural pattern of credential exposure. \texttt{KeywordDetector}
is disabled by default to avoid false positives on legitimate prose mentions
of words like \emph{key} or \emph{secret}, and can be enabled via
\texttt{code\_corpus\_mode=True} for code corpora. Rejected samples carry a
structured rejection reason listing all triggered detector types, with
per-type finding counts written to the provenance record.

\medskip
\begin{codeblock}[]
CuratorConfig(
    dataset="data/raw.jsonl",
    secrets_gate=True,
    secrets_code_corpus_mode=False,    # True for code corpora
)
\end{codeblock}

\subsection{PIIPseudonymizer}

\texttt{PIIPseudonymizer} replaces personally identifiable information with
consistent fake values, following the clinical NLP de-identification approach
used in i2b2 and MIMIC-III. The design pseudonymises rather than masks:
replacing \emph{John Smith} with \texttt{[PERSON]} corrupts document structure
and degrades QA generation quality, whereas a realistic fake name preserves
semantic coherence for all downstream generation and verification tasks.
Detection uses Microsoft Presidio with spaCy NER, and replacement values are
generated by Faker. Consistency is maintained through a per-sample entity map,
so the same entity string receives the same fake value across all fields within
a sample, with the map reset between samples to prevent cross-sample linkage.
A \texttt{faker\_seed} parameter makes replacements reproducible across runs.
The default entity type set covers \texttt{PERSON}, \texttt{EMAIL\_ADDRESS},
\texttt{PHONE\_NUMBER}, \texttt{US\_SSN}, \texttt{CREDIT\_CARD},
\texttt{IP\_ADDRESS}, \texttt{US\_BANK\_NUMBER}, and \texttt{IBAN\_CODE},
deliberately excluding \texttt{DATE\_TIME} to avoid over-redacting contract
metadata where dates carry semantic weight. The \texttt{ENTITY\_TYPES\_CLINICAL}
preset extends the default set with \texttt{DATE\_TIME}, \texttt{LOCATION},
and \texttt{MEDICAL\_LICENSE} for clinical or legal corpora where these
fields also constitute PII.

\medskip
\begin{codeblock}[]
from curatorkit.hygiene.pii import ENTITY_TYPES_CLINICAL

CuratorConfig(
    dataset="data/clinical_notes.jsonl",
    pii_pseudonymize=True,
    pii_entity_types=ENTITY_TYPES_CLINICAL,    # adds DATE_TIME, LOCATION, etc.
    pii_score_threshold=0.7,                   # Presidio confidence threshold
    pii_spacy_model="en_core_web_lg",          # or "en_core_web_sm" for dev/CI
    pii_faker_seed=42,                         # reproducible replacements
)
\end{codeblock}

\subsection{ToxicityGate}

\texttt{ToxicityGate} rejects samples with toxic content using a two-phase
approach that separates clear cases from ambiguous ones. In the first phase,
the Detoxify classifier~\cite{Detoxify} scores each sample across six toxicity
dimensions. Samples whose maximum dimension score falls below the pass
threshold are accepted immediately without any LLM call. Samples whose maximum
score exceeds the reject threshold are rejected immediately. Samples in the
intermediate range are escalated to an optional LLM judge for a second opinion,
incurring an LLM call only for genuinely ambiguous cases. This design avoids
paying for LLM inference on the large majority of samples that are
unambiguously clean or unambiguously toxic.
Three Detoxify model variants are supported: \texttt{"unbiased"} (default,
recommended for academic and professional corpora where sensitive topics appear
in legitimate context), \texttt{"original"} (for informal web text), and
\texttt{"multilingual"} for non-English datasets. For corpora where the subject
matter includes clinical, legal, or social science content, raising the pass
threshold to 0.2 avoids excessive escalation on text that discusses sensitive
topics without being toxic.

\medskip
\begin{codeblock}[]
CuratorConfig(
    dataset="data/raw.jsonl",
    toxicity_gate=True,
    toxicity_classifier_pass_threshold=0.1,
    toxicity_classifier_reject_threshold=0.5,
    toxicity_detoxify_model="unbiased",
    toxicity_llm_judge=True,               # escalate borderline samples
    toxicity_llm_reject_threshold=0.5,
    llm_model="openai/gpt-4o-mini",        # required when llm_judge=True
)
\end{codeblock}

\noindent All three hygiene components can be composed in a single
\texttt{CuratorConfig} invocation alongside a generation task. When all three
are active, they execute in the fixed cost-ordered sequence before any
generation call is made:

\medskip
\begin{codeblock}[]
CuratorConfig(
    dataset="data/raw.jsonl",
    secrets_gate=True,
    pii_pseudonymize=True,
    toxicity_gate=True,
    generation_task="qa",
    llm_model="openai/gpt-4o-mini",
    export_formats=["alpaca"],
    output_dir="output/",
)
\end{codeblock}

\section{Synthetic Data Generation}
\label{sec:generation}

The synthetic data generation layer takes normalised source samples and
produces training-ready outputs by issuing structured LLM calls, parsing
the responses, and stamping each output with a full provenance record. All
eight generation tasks extend \texttt{BaseGenerationTask}, which provides
synchronous and asynchronous run loops with \texttt{asyncio.Semaphore}
concurrency control, parse-failure retry, and rejected-sample accumulation
via \texttt{flush\_rejected()}. Parse failures are never dropped silently:
a sample whose LLM response cannot be parsed is emitted as a
\texttt{RejectedSample} with \texttt{rejection\_reason=generation\_parse\_failed},
and the raw LLM response is preserved in the provenance record for diagnosis.
All tasks are corpus-aware: when the input sample carries source text, it is
embedded in the generation prompt and stored in \texttt{sample.input} so
downstream hallucination verification can check the answer against the exact
evidence that produced it.

\subsection{LLM Backends}

CuratorKIT abstracts LLM access behind a \texttt{BaseLLM} interface with two
concrete backends. \texttt{LiteLLMBackend} routes through
LiteLLM~\cite{litellm2024} to 100+ providers including OpenAI, Anthropic, Google
Gemini, AWS Bedrock, Azure, vLLM~\cite{vllm}, and SGLang. It supports
\texttt{extra\_body} for provider-specific parameters, such as
\texttt{chat\_template\_kwargs} for thinking-mode control on Qwen3~\cite{qwen3_techreport}
and other reasoning models, and automatically strips \texttt{<think>} blocks
from reasoning model outputs. \texttt{OllamaBackend} provides a direct HTTP
client for local Ollama models, requiring no API key or external service.
Both backends expose synchronous \texttt{generate()} and asynchronous
\texttt{agenerate()} methods with built-in exponential backoff retry, token
tracking, and latency instrumentation. \texttt{LLMResponse} carries the
generated text, model identifier, token counts, and a
\texttt{to\_provenance\_dict()} method that serialises these fields directly
into the \texttt{ProvenanceRecord.notes} dictionary.
Separate generator and judge models are supported through the
\texttt{judge\_llm\_model} parameter, allowing a stronger model to evaluate
outputs without the same-model leniency bias that arises when a generator
scores its own responses~\cite{yuan2024selfrewarding,lee2023rlaif}.

\medskip
\begin{codeblock}[]
CuratorConfig(
    llm_model="openai/gpt-4o-mini",       # generator
    llm_temperature=0.7,
    llm_concurrency=16,
    judge_llm_model="openai/gpt-4o",      # judge -- stronger, separate model
    judge_llm_temperature=0.1,
    judge_concurrency=8,
)
\end{codeblock}

\subsection{Generation Tasks}

Eight generation tasks are available, summarised in
Table~\ref{tab:generation_tasks}. Each task owns only the prompt template
and the output parser; the LLM backend is injected at construction time,
making tasks composable with either backend without modification.

\begin{table}[h]
\small
\raggedright
\caption{Generation tasks available in CuratorKIT.}
\label{tab:generation_tasks}
\begin{tabular}{llp{6cm}}
\toprule
\textbf{Task} & \textbf{Output type} & \textbf{Description} \\
\midrule
\texttt{QAGenerationTask}          & \texttt{instruction\_following} & Grounded QA pairs per source chunk with difficulty control \\
\texttt{PreferenceGenerationTask}  & \texttt{preference}             & Chosen/rejected pairs for DPO; single-call or two-pass mode \\
\texttt{GRPORolloutTask}           & \texttt{grpo}                   & $K$ temperature-varied responses per prompt with optional scoring \\
\texttt{MultiTurnTask}             & \texttt{conversational}         & Multi-turn dialogues via turn-by-turn or single-call generation \\
\texttt{EvolInstructTask}          & \texttt{instruction\_following} & Instruction rewrites across five complexity strategies \\
\texttt{ChainOfThoughtTask}        & \texttt{instruction\_following} & Step-by-step reasoning in generate or wrap mode \\
\texttt{AdversarialQAGenerationTask} & \texttt{instruction\_following} & QA pairs with controlled hallucination injection for gate evaluation \\
\texttt{BadSampleInjector}         & \texttt{instruction\_following} & Post-hoc adversarial injection utility for gate benchmarking \\
\bottomrule
\end{tabular}
\end{table}

\textbf{QAGenerationTask} is the primary text-to-SFT task. It generates $N$
question-answer pairs per source chunk with configurable difficulty
(\texttt{easy}, \texttt{medium}, \texttt{hard}) and separate table-specific
prompts for chunks extracted from PDF tables. A \texttt{run\_multi\_passage()}
mode generates cross-passage questions that require information from two
adjacent chunks, producing harder samples less susceptible to single-chunk
memorisation.

\begin{codeblock}[]
CuratorConfig(
    dataset="docs/handbook.pdf",
    llm_model="openai/gpt-4o-mini",
    generation_task="qa",
    num_questions=3,
    difficulty="medium",
)
\end{codeblock}

\medskip

\textbf{PreferenceGenerationTask} produces chosen/rejected pairs for DPO
training~\cite{rafailov2023direct}. In \texttt{single\_call} mode, one LLM call produces
both responses. In \texttt{two\_pass} mode, the chosen response is generated
first and the rejected response is generated separately at higher temperature,
achieving stronger quality contrast between the two completions.

\medskip

\textbf{GRPORolloutTask} generates $K$ diverse responses per prompt for
group relative policy optimisation~\cite{shao2024deepseekmath}. Temperatures across the $K$
rollouts are either specified explicitly or derived from a configurable spread
around the base temperature, ensuring sufficient diversity in the response
set. Rollouts can be scored by a separate LLM judge in the same pipeline step.

\begin{codeblock}[]
CuratorConfig(
    dataset="data/prompts.jsonl",
    llm_model="openai/gpt-4o-mini",
    generation_task="grpo",
    num_responses=6,
    grpo_temperatures=[0.0, 0.3, 0.6, 0.9, 1.2, 1.5],
    score_responses=True,
)
\end{codeblock}

\medskip

\textbf{MultiTurnTask} generates conversational data. In \texttt{turn\_by\_turn}
mode (default), each turn is an independent LLM call conditioned on all prior
turns, mirroring how real RLHF conversation data is collected and avoiding
the distribution collapse of single-call generation. This uses $2 \times
\texttt{num\_turns}$ LLM calls per sample.

\textbf{EvolInstructTask} rewrites instructions using five complexity
strategies: \texttt{add\_constraints}, \texttt{deepen}, \texttt{concretize},
\texttt{increase\_reasoning}, and \texttt{broaden}, following the
Evol-Instruct paradigm~\cite{xu2023wizardlm}. When source text is present,
evolutions remain grounded to the corpus. Answers can be generated via an
optional second LLM pass.

\textbf{ChainOfThoughtTask} operates in two modes: \texttt{generate} produces
a full reasoning chain from scratch, and \texttt{wrap} takes an existing
instruction-output pair and generates the reasoning chain leading to that
answer~\cite{wang2023selfconsistency}.

\textbf{AdversarialQAGenerationTask} extends \texttt{QAGenerationTask} to
inject controlled hallucinations at a configurable rate during the first
generation pass. Five injection types are supported:
\emph{contradicts\_source}, \emph{parametric\_drift},
\emph{high\_temperature\_drift}, \emph{domain\_mismatch}, and
\emph{instruction\_quality}. Since the hallucination gate evaluates these
samples without access to the injection metadata, this provides ground-truth
labels for gate precision and recall measurement without requiring human
annotation. 

\textbf{BadSampleInjector} is a complementary post-hoc utility
that takes already-generated samples and replaces a fraction of answers with
adversarially hallucinated variants. It is intended for gate evaluation
experiments rather than standard pipeline use.

\subsection{Custom Prompt Templates}

Every generation task and quality gate accepts a prompt template override.
Templates are validated at construction time against a list of required
placeholder variables, so a missing \texttt{\{context\}} or
\texttt{\{num\_questions\}} raises a \texttt{ValueError} with the full
variable list before any LLM call is made.

\medskip
\begin{codeblock}[]
CuratorConfig(
    generation_task="qa",
    qa_prompt_template=(
        "You are a legal analyst. Generate {num_questions} precise "
        "{difficulty} questions from the clause below. Each question "
        "must be answerable from the text alone.\n\n"
        "Clause:\n{context}\n\n"
        "Return JSON: [{\"question\": \"...\", \"answer\": \"...\"}]"
    ),
)
\end{codeblock}

\section{Quality Filtering \& Adaptive Recovery}
\label{sec:filtering}

Generated samples pass through a sequence of three quality gates before
export. Each gate operates on a distinct signal axis: source grounding,
instruction-following quality, and semantic diversity. Samples that fail a
gate are not discarded immediately. Every rejection triggers a structured
diagnostic process that classifies the failure into one of nine categories
and, where the failure is recoverable, applies a targeted repair rather than
blind resampling. This section describes the three gates, the failure
taxonomy, and the recovery system.

\subsection{Hallucination Gate}

\texttt{HallucinationGate} uses an LLM judge to verify that each generated
answer is grounded in its source text. The judge evaluates four dimensions:
factual accuracy, source grounding, non-contradiction, and absence of
parametric drift. It produces a \texttt{grounding\_score} in $[0, 1]$ along
with lists of supported and unsupported claims. Samples below the configured
threshold (default 0.7) are rejected with a structured reason of the form
\texttt{hallucination\_contract\_failed:0.42}.
The verification is provenance-exact: the judge receives the source chunk
from \texttt{sample.input}, which is the same text stored by the generation
task at the time of generation. This eliminates the retrieval error class that
affects post-hoc grounding checks, where the retrieved chunk may be similar
to but not identical to the original generation context~\cite{maynez2020faithfulness,rashkin2021increasing}.
For DPO preference pairs, only the chosen response is evaluated. For GRPO
rollouts, the highest-scoring response is checked. Using the same model for
both generation and judgement introduces same-model leniency bias~\cite{yuan2024selfrewarding};
a separate, typically stronger judge model is recommended.

\medskip
\begin{codeblock}[]
CuratorConfig(
    hallucination_threshold=0.7,
    judge_llm_model="openai/gpt-4o",    # recommended: separate from generator
    judge_llm_temperature=0.1,
)
\end{codeblock}

\subsection{Reward Gate}

\texttt{RewardGate} scores samples along configurable quality dimensions
following the UltraFeedback rubric~\cite{cui2023ultrafeedback}. Seven dimensions are
available: \texttt{helpfulness}, \texttt{honesty}, \texttt{instruction\_following},
\texttt{truthfulness}, \texttt{depth}, \texttt{creativity}, and
\texttt{coherence}. The judge produces an overall score as the mean of
all dimension scores, alongside a per-dimension breakdown and free-text
strengths and weaknesses. The per-dimension breakdown is consumed directly
by the \texttt{RewardRefiner} during recovery.
For DPO preference pairs, the gate applies dual scoring: the chosen response
must clear the threshold and the rejected response must fall below it. A pair
where the rejected response also clears the threshold is dropped for
insufficient quality contrast, as such pairs provide a weak training signal
for preference learning. Contrast failures of this kind are not forwarded to
the recovery system, since rewriting the answer cannot resolve a failure in
the generation contrast itself.

\medskip
\begin{codeblock}[]
CuratorConfig(
    reward_threshold=0.7,
    reward_dimensions=["helpfulness", "honesty", "instruction_following", "depth"],
    judge_llm_model="openai/gpt-4o",
)
\end{codeblock}

\subsection{Diversity Gate}

\texttt{DiversityGate} filters near-duplicate samples via embedding cosine
similarity using sentence-transformers~\cite{sentence-transformers}. Accepted samples form a
growing reference set within the current pipeline run; each new candidate is
compared against all accepted samples and rejected if its maximum similarity
exceeds the threshold (default 0.92). FAISS~\cite{douze2025faisslibrary} is used for
approximate nearest neighbour search when available, with a NumPy brute-force
fallback otherwise. The text field to embed is selected automatically from
\texttt{task\_type}, using \texttt{chosen} for preference pairs,
\texttt{instruction} for GRPO rollouts, and \texttt{instruction + output} for
SFT samples.

\medskip
\begin{codeblock}[]
CuratorConfig(
    diversity_threshold=0.92,
    embedding_model="sentence-transformers/all-MiniLM-L6-v2",
    embedding_device=None,    # "cuda" | "cpu" | None (auto-detect)
)
\end{codeblock}

\subsection{Failure Mode Taxonomy}

Every gate rejection enters a structured diagnostic process rather than being
discarded. The diagnostic system classifies each rejection into one of nine
failure modes, six of which are recoverable through targeted repair. The
taxonomy is organised by the gate that produced the rejection, since each
gate surfaces a distinct class of failure and each class admits a different
repair strategy.

\begin{table}[h]
\small
\raggedright
\caption{Failure mode taxonomy. Recoverable modes are addressed by the
inline probe and reward refiner.}
\label{tab:failure_modes}
\begin{tabular}{llp{5.0cm}c}
\toprule
\textbf{Origin} & \textbf{Mode} & \textbf{Description} & \textbf{Recoverable} \\
\midrule
\multirow{4}{*}{HallucinationGate}
  & \texttt{THRESHOLD\_MARGINAL}    & Score just below threshold; outcome unstable at current temperature  & Yes \\[4pt]
  & \texttt{GENERATOR\_TEMPERATURE} & High temperature caused source drift                                 & Yes \\[4pt]
  & \texttt{GENERATOR\_PARAMETRIC}  & Model ignored source and relied on prior knowledge                  & Yes \\[4pt]
  & \texttt{SOURCE\_AMBIGUOUS}      & Source-response relationship is fundamentally unclear               & No  \\
\midrule
\multirow{3}{*}{RewardGate}
  & \texttt{INSTRUCTION\_QUALITY}   & Generated question is poor or ambiguous                             & Yes \\[4pt]
  & \texttt{RESPONSE\_QUALITY}      & Generated answer is weak or incoherent                              & Yes \\[4pt]
  & \texttt{DOMAIN\_MISMATCH}       & Generation prompt ill-suited to domain style                        & Yes \\
\midrule
DiversityGate
  & \texttt{NEAR\_DUPLICATE}        & Sample too similar to an already-accepted sample                    & No  \\
\midrule
Fallback
  & \texttt{UNKNOWN}                & Probe sequence exhausted without a conclusive diagnosis             & No  \\
\bottomrule
\end{tabular}
\end{table}

\subsection{Diagnostic Probe}

\texttt{DiagnosticProbe} runs inline on every \texttt{HallucinationGate}
rejection, consuming at most five LLM calls per sample across two routing
paths. The routing decision is conditioned on the grounding score written to
the sample's provenance record by the failing gate evaluation.
For samples whose grounding score is at or above the \texttt{score\_split}
threshold (default 0.5), a near-boundary failure is assumed and the probe
runs a temperature sweep first, regenerating the sample at each configured
temperature (default $T \in \{0.3, 0.5\}$). If all regenerations pass the
gate, the mode is classified as \emph{THRESHOLD\_MARGINAL}; if only the
lower temperature passes, it is classified as \emph{GENERATOR\_TEMPERATURE}.
If the sweep is inconclusive, the probe proceeds to the prompt variant
sequence.
For samples scoring below \texttt{score\_split}, a deeper source grounding
failure is assumed and the strict grounding probe is attempted first. A pass
on this path classifies the mode as \emph{GENERATOR\_PARAMETRIC}, confirming
that the model was relying on prior knowledge rather than the provided source
passage. If the strict grounding probe fails, the temperature sweep is run,
followed by the remaining prompt variants.
The prompt variant sequence covers three cases in order: a strict grounding
prompt targeting \emph{GENERATOR\_PARAMETRIC}, a domain-adapted prompt
targeting \emph{DOMAIN\_MISMATCH}, and an instruction regeneration step
targeting \emph{INSTRUCTION\_QUALITY}, where the probe rewrites the question
from the source rather than the answer. The first regeneration that clears
the gate is accepted and reintegrated into the pipeline inline, flowing into
\texttt{RewardGate} alongside the original passing samples. If all five
probe calls are exhausted without recovery, the sample is classified as
\emph{SOURCE\_AMBIGUOUS} or \emph{UNKNOWN} and written to the rejection log.

\medskip
\begin{codeblock}[]
CuratorConfig(
    enable_diagnostic_probe=True,
    probe_temperatures=[0.3, 0.5],
    probe_score_split=0.5,
    probe_generator_model=None,    # None = use llm_model
)
\end{codeblock}

Probe templates can be overridden for domain-specific recovery strategies.
Each template receives \texttt{\{source\}} and \texttt{\{question\}} as
placeholders:

\medskip
\begin{codeblock}[]
CuratorConfig(
    enable_diagnostic_probe=True,
    probe_extra_templates={
        "strict_grounding": (
            "Answer using ONLY the passage. Quote specific sentences.\n\n"
            "Passage:\n{source}\n\nQuestion:\n{question}"
        ),
        "domain_specific": (
            "You are a legal analyst. Answer from the passage only.\n\n"
            "Passage:\n{source}\n\nQuestion:\n{question}"
        ),
    },
)
\end{codeblock}

\subsection{Reward Refiner}

\texttt{RewardRefiner} operates on samples that fail \texttt{RewardGate}
after the inline probe has been exhausted. It reads the lowest-scoring
dimension and the judge's associated weakness note from the gate's provenance
record, then issues a single targeted rewrite prompting the generator to
improve that specific axis while remaining grounded in the source passage.
The refined sample is re-evaluated by the same \texttt{RewardGate} instance.
If it passes, it joins the accepted pool. For DPO pairs, the refined answer
becomes the new \texttt{chosen} field while the original adversarial
\texttt{rejected} response is preserved unchanged, maintaining the pair
structure for downstream preference training.
Samples rejected for insufficient DPO quality contrast
(\texttt{rejected\_above\_threshold}) are skipped entirely by the refiner,
as the failure lies in the generation contrast rather than the answer quality
and cannot be resolved by rewriting.

\medskip
\begin{codeblock}[]
CuratorConfig(
    enable_reward_refiner=True,
    reward_refine_prompt_template=None,
    reward_instruction_refine_template=None,
)
\end{codeblock}

\subsection{Pipeline Diagnostics}

\texttt{PipelineDiagnostics} is a run-level accumulator that collects every
diagnosed rejection during a pipeline run. It is accessible as
\texttt{result.diagnostics} after \texttt{Curator.run()} completes and
writes a \texttt{diagnostic\_summary.json} to the output directory alongside
the standard manifest files.

\medskip
\begin{codeblock}[]
result = curator.run()
if result.diagnostics:
    d = result.diagnostics
    print(f"Diagnosed  : {d.to_dict()['total_diagnosed']}")
    print(f"Recovered  : {d.probe_recovery_count()}")
    print(f"Probe calls: {d.total_probe_calls()}")
    print(f"Mode counts: {d.mode_counts()}")
\end{codeblock}

\section{Usage}
\label{sec:usage}

\subsection{Installation}

\begin{codeblock}[listing options={language=bash}]
git clone https://github.com/Lexsi-Labs/CuratorKIT.git
cd CuratorKIT
pip install -e .                         # core: cleaning + dedup only
pip install -e ".[generation]"           # + LLM backends (LiteLLM)
pip install -e ".[connectors]"           # + Parquet + HuggingFace readers
pip install -e ".[embedding]"            # + diversity gate + cross-run dedup
pip install -e ".[embedding-faiss]"      # + FAISS-accelerated diversity gate
pip install -e ".[hygiene]"              # + SecretsGate + PIIPseudonymizer + ToxicityGate
pip install -e ".[pdf]"                  # + PDF extraction (GPU auto-detected)
pip install -e ".[all]"                  # connectors + generation + embedding + hygiene
\end{codeblock}

\subsection{Usage Patterns}

\textbf{Pattern 1: Clean and deduplicate an existing dataset.} No LLM
required. Reads any supported format, deduplicates, cleans, and exports:

\begin{codeblock}[]
from curatorkit import Curator, CuratorConfig

result = Curator(CuratorConfig(
    dataset="Anthropic/hh-rlhf",
    dedup="minhash",
    clean=True,
    export_formats=["alpaca", "sharegpt"],
    output_dir="output/clean",
)).run()

result.print_summary()
# --------------------------------------------
#   passed   :      8 120
#   rejected :        340
#   time     :      12.3s
#   output   : output/clean
# --------------------------------------------
\end{codeblock}

\clearpage
\textbf{Pattern 2: Screen source data for hygiene before generation.}
Runs the three hygiene gates on the input corpus before any LLM API call is
made. API keys and PII in source data never reach an external endpoint:

\begin{codeblock}[]
result = Curator(CuratorConfig(
    dataset="data/corpus.jsonl",
    secrets_gate=True,
    secrets_hex_limit=4.5,          # raise from 3.0 for prose corpora
    secrets_base64_limit=5.5,       # raise from 4.5 for prose corpora
    pii_pseudonymize=True,
    toxicity_gate=True,
    export_formats=["alpaca"],
    output_dir="output/hygiene",
)).run()
\end{codeblock}
\vspace{10pt}
Each gate is independent: \texttt{SecretsGate} rejects samples containing
credentials or high-entropy tokens; \texttt{PIIPseudonymizer} replaces
detected entities in-place using Presidio NER and Faker; \texttt{ToxicityGate}
rejects samples whose Detoxify score exceeds the rejection threshold.
Rejected samples are written to \texttt{rejected.jsonl} with structured
rejection reasons for post-run audit.

\vspace{10pt}
\textbf{Pattern 3: Generate synthetic data and filter it.} Reads a PDF,
generates QA pairs, and checks that each answer is grounded in its source chunk:

\begin{codeblock}[]
result = Curator(CuratorConfig(
    dataset="docs/handbook.pdf",
    llm_model="openai/gpt-4o-mini",
    generation_task="qa",
    num_questions=3,
    hallucination_threshold=0.7,
    export_formats=["alpaca"],
    output_dir="output/qa",
)).run()
\end{codeblock}

\vspace{10pt}
\textbf{Pattern 4: Generate, filter, and recover failures.} Adds a reward
gate, inline diagnostic probe (fires on both \texttt{HallucinationGate} and
\texttt{RewardGate} rejections), and a post-pipeline reward refiner:

\begin{codeblock}[]
result = Curator(CuratorConfig(
    dataset="docs/handbook.pdf",
    llm_model="openai/gpt-4o-mini",
    judge_llm_model="openai/gpt-4o",      # separate judge avoids self-leniency
    generation_task="qa",
    hallucination_threshold=0.7,
    reward_threshold=0.7,
    enable_diagnostic_probe=True,
    enable_reward_refiner=True,
    export_formats=["alpaca", "sharegpt", "dpo"],
    output_dir="output/qa_full",
)).run()

if result.diagnostics:
    d = result.diagnostics
    print(f"Diagnosed  : {d.to_dict()['total_diagnosed']}")
    print(f"Recovered  : {d.probe_recovery_count()}")
    print(f"Probe calls: {d.total_probe_calls()}")
\end{codeblock}

\subsection{Custom Generation Tasks and Gates}

New generation tasks subclass \texttt{BaseGenerationTask} and implement two
methods. New gates subclass \texttt{BaseGate} and implement \texttt{run()}:

\begin{codeblock}[]
from curatorkit.generators.base import BaseGenerationTask
from curatorkit.llm.base import BaseLLM, LLMResponse
from curatorkit.schema import DataSample
import uuid

class SummarisationTask(BaseGenerationTask):
    def _build_messages(self, sample: DataSample) -> list[dict]:
        source = self._get_source_context(sample)
        return [{"role": "user", "content": f"Summarise:\n\n{source}"}]

    def _parse_response(self, sample: DataSample, response: LLMResponse):
        text = response.text.strip()
        if not text:
            return []
        return [DataSample(
            id=str(uuid.uuid4()),
            source_uri=sample.source_uri,
            instruction="Summarise the passage.",
            input=self._get_source_context(sample),
            output=text,
            task_type="instruction_following",
            provenance_chain=list(sample.provenance_chain),
        )]
\end{codeblock}

\medskip

Use custom tasks directly with \texttt{Pipeline} (bypassing
\texttt{CuratorConfig}):

\begin{codeblock}[]
from curatorkit.pipeline import Pipeline
from curatorkit.llm.litellm import LiteLLMBackend

llm  = LiteLLMBackend(model="openai/gpt-4o-mini")
task = SummarisationTask(llm=llm, concurrency=10)

pipeline = Pipeline([reader, schema_gate, task, alpaca_exporter],
                    output_dir=Path("output/"))
result = pipeline.run()
\end{codeblock}

\subsection{PDF Setup}

MinerU model weights are downloaded automatically on first use, but the
installation can be verified explicitly:

\begin{codeblock}[listing options={language=bash}]
curatorkit setup-pdf --check             # verify MinerU installation
\end{codeblock}

\subsection{CLI and YAML Configuration}

The CLI entry point is \texttt{curatorkit}. Pipelines can be defined entirely
in YAML, validated by Pydantic before any step runs:

\begin{codeblock}[listing options={language=yaml}]
# pipeline.yaml
name: "qa-pipeline"
version: "0.2.0"

readers:
  - type: pdf
    path: "corpus/contracts.pdf"
    chunk_strategy: heading
    extract_tables: true

gates:
  - type: schema
    min_tokens: 20
    max_tokens: 4096

  - type: secrets
    secrets_code_corpus_mode: false
    secrets_hex_limit: 4.5
    secrets_base64_limit: 5.5

  - type: hallucination
    hallucination_threshold: 0.7

  - type: reward
    reward_threshold: 0.6
    reward_dimensions: [helpfulness, honesty, instruction_following]

  - type: toxicity
    toxicity_classifier_pass_threshold: 0.1
    toxicity_classifier_reject_threshold: 0.5

normalizers:
  - type: exact_dedup
  - type: minhash_dedup
    minhash_threshold: 0.85
  - type: text_cleaner
  - type: pii_pseudonymizer
    pii_score_threshold: 0.7

generators:
  - type: qa
    num_questions: 3
    difficulty: medium

exporters:
  - type: alpaca
  - type: sharegpt
  - type: dpo

llm:
  model: openai/gpt-4o-mini
  temperature: 0.7
  concurrency: 16

diagnostic:
  enable_probe: true
  probe_temperatures: [0.3, 0.5]

output_split:
  train: 0.8
  val: 0.1
  test: 0.1
\end{codeblock}

\begin{codeblock}[listing options={language=bash}]
# Validate config and print execution plan -- no LLM calls
curatorkit run pipeline.yaml --dry-run

# Full run
curatorkit run pipeline.yaml --output-dir ./output/ --verbose

# Async execution for generation-heavy pipelines
curatorkit run pipeline.yaml --async --verbose

# Reset cross-run dedup index before running
curatorkit run pipeline.yaml --reset-index
\end{codeblock}

\medskip

\textbf{YAML for local vLLM endpoints:}

\begin{codeblock}[listing options={language=yaml}]
llm:
  model: openai/Qwen/Qwen3-8B
  api_base: http://localhost:8000/v1
  api_key: token-abc123
  temperature: 0.7
  concurrency: 32
  extra_body:
    chat_template_kwargs:
      enable_thinking: false
\end{codeblock}

\subsection{Custom Prompt Templates}

Every generation task and gate accepts a \texttt{*\_prompt\_template}
override. Templates are validated at construction time: if a required
placeholder is absent, a \texttt{ValueError} listing all required variables
is raised before any LLM call is made:

\begin{codeblock}[]
CuratorConfig(
    generation_task="qa",
    qa_prompt_template=(
        "You are a legal analyst. Generate {num_questions} precise "
        "{difficulty} questions from the clause below. Each question "
        "must be answerable from the text alone.\n\n"
        "Clause:\n{context}\n\n"
        "Return JSON: [{\"question\": \"...\", \"answer\": \"...\"}]"
    ),
)
\end{codeblock}

\medskip

The reward judge prompt can be replaced entirely to define a custom evaluation
rubric. The prompt receives \texttt{\{instruction\}} and \texttt{\{response\}}
as placeholders and must return a JSON object containing at minimum an
\texttt{overall\_score} field in $[0, 1]$:

\begin{codeblock}[]
CuratorConfig(
    reward_threshold=0.7,
    reward_prompt_template=(
        "Rate this legal answer 0.0-1.0:\n"
        "- Cites clause/article (0.4 weight)\n"
        "- States obligation precisely (0.3 weight)\n"
        "- Identifies exceptions (0.3 weight)\n\n"
        "Instruction: {instruction}\nResponse: {response}\n\n"
        "Return JSON: {\"overall_score\": 0.XX, \"reasoning\": \"...\"}"
    ),
)
\end{codeblock}

\section{Feature Comparison}
\label{sec:comparison}

Table~\ref{tab:comparison} compares CuratorKIT against related data curation
frameworks.

\begin{table}[ht]
  \centering
  \small
  \caption{Feature comparison across data curation frameworks.
    \cmark = full support, \tmark = partial, \xmark = not available.}
  \label{tab:comparison}
  \begin{tabularx}{\textwidth}{>{\raggedright\arraybackslash}p{0.28\textwidth}CCCCC}
    \toprule
    \rowcolor{black!4}
    & \textbf{Distilabel} & \textbf{DataTrove} & \textbf{NeMo DD} & \textbf{AgentInstruct} & \textbf{CuratorKIT} \\
    \midrule
    \multicolumn{6}{l}{\textbf{Ingestion}} \\[1pt]
    \quad Multi-format readers ($\ge$5) & \cmark & \cmark & \tmark & \xmark & \cmark \\[4pt]
    \multicolumn{6}{l}{\textbf{Generation}} \\[1pt]
    \quad Generation tasks ($\ge$5) & \cmark & \tmark & \cmark\textsuperscript{4} & \cmark & \cmark\textsuperscript{5} \\[4pt]
    \multicolumn{6}{l}{\textbf{Quality enforcement}} \\[1pt]
    \quad Schema validation & \xmark & \xmark & \cmark & \xmark & \cmark \\
    \quad Hallucination / faithfulness gate & \tmark\textsuperscript{1} & \xmark & \tmark\textsuperscript{2} & \xmark & \cmark \\
    \quad Reward / quality scoring gate & \cmark & \xmark & \tmark & \xmark & \cmark \\
    \quad Diversity / semantic dedup gate & \cmark & \xmark & \xmark & \xmark & \cmark \\
    \quad Deduplication (exact + fuzzy + cross-run) & \cmark & \cmark & \xmark & \xmark & \cmark \\[4pt]
    \multicolumn{6}{l}{\textbf{Provenance \& observability}} \\[1pt]
    \quad Per-sample provenance chain & \tmark\textsuperscript{3} & \xmark & \xmark & \xmark & \cmark \\
    \quad Structured rejection reasons & \xmark & \tmark & \xmark & \xmark & \cmark \\[4pt]
    \multicolumn{6}{l}{\textbf{Recovery}} \\[1pt]
    \quad Adaptive recovery (diagnose + repair) & \xmark & \xmark & \xmark & \xmark & \cmark \\[4pt]
    \multicolumn{6}{l}{\textbf{Usability}} \\[1pt]
    \quad YAML + CLI + Python API & \cmark & \tmark & \cmark & \xmark & \cmark \\
    \bottomrule
  \end{tabularx}

  \vspace{3pt}
  \begin{minipage}{\textwidth}
  \scriptsize\color{black!60}
  \textsuperscript{1}~UltraFeedback truthfulness; no source-grounded verification.\quad
  \textsuperscript{2}~LLM judge accuracy rubric; no per-sample source grounding.\quad
  \textsuperscript{3}~Pipeline-level YAML lineage; no per-row trace. 
  \textsuperscript{4}~NVIDIA NIM endpoints only.\quad
  \textsuperscript{5}~Via LiteLLM plus native Ollama backend.
  \end{minipage}
\end{table}

Key structural differences from other frameworks:

\begin{itemize}
  \item \textbf{Rejection as a structured event.} Every rejected sample
    carries a machine-readable reason string, the rejecting step identity,
    and the configuration snapshot. Without a structured failure reason, the
    only actionable response to a gate failure is discard. With one, the
    recovery system can route the sample to a targeted repair rather than
    treating all failures as equivalent.

  \item \textbf{Rejection as the start of recovery.} Other frameworks treat
    quality gates as terminal. CuratorKIT's diagnostic loop classifies each
    failure and applies a targeted repair: different interventions for
    temperature-driven drift, parametric leakage, instruction quality issues,
    and source ambiguity. Blind resampling applies the same intervention
    regardless of cause; targeted repair applies the right one.

  \item \textbf{Provenance without retrieval.} Hallucination verification
    requires the verifier to see the same evidence the generator saw.
    Retrieval-based approaches approximate this by re-fetching a similar
    chunk, introducing error when the returned chunk is close but not
    identical to the original. Storing the source chunk in the sample's own
    fields eliminates this error class at no infrastructure cost.

  \item \textbf{Separate generator and judge models.} The architecture
    explicitly supports different models for generation and evaluation,
    avoiding same-model leniency bias documented in self-rewarding
    setups~\cite{yuan2024selfrewarding}.
\end{itemize}

\section{Empirical Studies}
\label{sec:empirical}

We validated CuratorKIT across three generator scales from the Qwen3
family~\cite{qwen3_techreport} (1.7B, 4B, 8B) and two judge scales (14B, 35B). The
source corpus comprises 1,500 chunks drawn from CUAD~\cite{cuad} (commercial
contracts), PubMedQA~\cite{jin2019pubmedqa} (biomedical abstracts), and Wikipedia,
with three QA pairs generated per chunk and 20\% of samples seeded with
controlled hallucination injections across four failure types:
\emph{contradicts\_source}, \emph{parametric\_drift}, \emph{domain\_mismatch},
and \emph{instruction\_quality}. The downstream fine-tuning evaluation uses
Qwen3-4B as the base model with LoRA (rank 16, $\alpha=32$), evaluated on a
frozen test set of 1,400 samples.

\subsection{Gate Complementarity}

The hallucination and reward gates operate on orthogonal signal axes and
reject largely disjoint failure populations. Table~\ref{tab:gate_overlap}
reports the Jaccard similarity between the sets of samples rejected by each
gate independently. Overlap stays in the range $[0.28, 0.32]$ across all
three generator scales, confirming that neither gate subsumes the other.
Running reward-only filtering misses the faithfulness failures caught
exclusively by the hallucination gate; running hallucination-only filtering
misses the quality failures caught exclusively by the reward gate.

\begin{table}[h]
\small
\centering
\caption{Rejection-set Jaccard overlap between \texttt{HallucinationGate}
and \texttt{RewardGate} across generator scales (14B judge).}
\label{tab:gate_overlap}
\begin{tabular}{lcccc}
\toprule
\textbf{Generator} & \textbf{Hall-only (\%)} & \textbf{Reward-only (\%)} & \textbf{Both (\%)} & \textbf{Jaccard} \\
\midrule
1.7B & 7.3 & 32.3 & 18.4 & 0.317 \\
4B   & 6.5 & 38.6 & 21.6 & 0.324 \\
8B   & 8.5 & 36.6 & 17.8 & 0.284 \\
\bottomrule
\end{tabular}
\end{table}

\subsection{Downstream Fine-Tuning Quality}

Table~\ref{tab:downstream_quality} compares unfiltered generation against
both-gates-filtered generation on ROUGE-L~\cite{lin2004rouge}, BERTScore-F1~\cite{zhang2020bertscore},
and faithfulness across all three generator scales. Every delta is strictly
positive across all three metrics at every generator scale: no metric
regresses under filtering. The gain is largest where the generator is weakest,
reaching $+3.1\%$ ROUGE-L at 1.7B, and compresses as generator quality
increases. Generator scale is the dominant driver of downstream quality,
with a cross-generator ROUGE-L improvement of $+6.6\%$ from 1.7B to 8B
dwarfing any within-generator spread across curation conditions.

\begin{table}[h]
\small
\centering
\caption{Downstream fine-tuning quality: unfiltered vs.\ both gates applied.
Qwen3-4B base model, LoRA, 1,400-sample test set.}
\label{tab:downstream_quality}
\begin{tabular}{llccc}
\toprule
\textbf{Generator} & \textbf{Condition} & \textbf{ROUGE-L} & \textbf{BERTScore-F1} & \textbf{Faithfulness} \\
\midrule
\multirow{3}{*}{1.7B}
  & Unfiltered & 0.5227 & 0.9059 & 0.8537 \\
  & Both gates & 0.5388 & 0.9112 & 0.8606 \\
  & $\Delta$   & \textbf{+0.0161 (+3.1\%)} & +0.0053 (+0.6\%) & +0.0069 (+0.8\%) \\
\midrule
\multirow{3}{*}{4B}
  & Unfiltered & 0.5577 & 0.9185 & 0.8542 \\
  & Both gates & 0.5606 & 0.9197 & 0.8570 \\
  & $\Delta$   & +0.0029 (+0.5\%) & +0.0012 (+0.1\%) & +0.0028 (+0.3\%) \\
\midrule
\multirow{3}{*}{8B}
  & Unfiltered & 0.5734 & 0.9187 & 0.8527 \\
  & Both gates & 0.5764 & 0.9191 & 0.8540 \\
  & $\Delta$   & +0.0030 (+0.5\%) & +0.0004 (+0.04\%) & +0.0013 (+0.2\%) \\
\bottomrule
\end{tabular}
\end{table}

\subsection{Adaptive Recovery}

The adaptive recovery pipeline (DiagnosticProbe and RewardRefiner) is
evaluated against hard filtering and against naive regeneration retry.
Compared to hard filtering at identical quality thresholds, adaptive recovery
yields $+23.6\%$ more accepted samples at 1.7B, $+48.4\%$ at 4B, and
$+53.1\%$ at 8B, with the yield advantage growing with generator scale.
Against naive retry under the same judge and thresholds, the adaptive
pipeline achieves a higher recovery rate (32.5\% vs.\ 24.9\% of all
rejections recovered) and substantially higher injection recall across all
four failure types (67.0\% vs.\ 53.5\% overall, with a 23 percentage point
gap on \emph{instruction\_quality} failures). Structured diagnosis routes
repairs to the failure type rather than issuing a blind regeneration from
the same prompt, which is the source of its advantage on systematically
diagnosable failures.

\begin{table}[h]
\small
\centering
\caption{Adaptive recovery vs.\ naive retry: injection recall by failure type
(Qwen3-1.7B generator, 14B judge).}
\label{tab:recovery_recall}
\begin{tabular}{lccc}
\toprule
\textbf{Injection type} & \textbf{Adaptive recall} & \textbf{Naive retry recall} & \textbf{Gap} \\
\midrule
\emph{contradicts\_source}  & 59.8\% & 49.4\% & +10.4pp \\
\emph{parametric\_drift}    & 71.7\% & 58.1\% & +13.6pp \\
\emph{domain\_mismatch}     & 65.3\% & 61.3\% & +4.0pp  \\
\emph{instruction\_quality} & 74.4\% & 51.3\% & +23.1pp \\
\textbf{Overall}            & \textbf{67.0\%} & \textbf{53.5\%} & \textbf{+13.5pp} \\
\bottomrule
\end{tabular}
\end{table}

\subsection{Data Hygiene Validation}

The hygiene layer was validated across two notebooks. Notebook 07 used
CuratorKIT's \texttt{qa\_prompt\_template} parameter to instruct a
Qwen3-8B generator to produce four structured contamination batches,
covering answers embedding API keys and tokens, first-person instructions
containing names, emails, and SSNs, hostile forum-style answers, and a
clean control batch. The four batches were combined into a single
184-sample contaminated dataset and passed to Notebook 08.
Notebook 08 ran the full hygiene pipeline on this input.
\texttt{SecretsGate} rejected all 184 samples before any reached the
downstream components. Since \texttt{SecretsGate} runs as the first and
cheapest gate in the fixed execution order, the entire contaminated
batch was intercepted before any PII normalisation or toxicity scoring
was invoked, and no credentials reached an external LLM endpoint. This
confirms that the cost-ordered gate sequence works as intended, with the
most structurally obvious contamination eliminated at zero model cost
before the more expensive components are ever invoked.

\section{Limitations}
\label{sec:limitations}

CuratorKIT has several practical limitations. The \texttt{DiagnosticProbe} can
issue up to five sequential LLM calls per rejected sample; although
\texttt{diagnose\_batch()} parallelizes probes across samples, per-sample
latency remains significant and should be considered when estimating pipeline
throughput. Recovery is currently limited to corpus-grounded generation tasks,
since the probe requires a populated \texttt{sample.input} field for
regeneration, and instruction-only or open-ended samples bypass this stage.
In addition, \texttt{RewardRefiner} performs only one targeted rewrite per
sample and does not continue iterating if the rewritten sample fails
re-evaluation. Some components require language-specific configuration such as
\texttt{PIIPseudonymizer} defaults to English Presidio models, so non-English
corpora require explicit spaCy model configuration and may have lower detection
accuracy. CuratorKIT currently operates on text only and does not support
multimodal pipelines. Finally, the nine-mode failure taxonomy used by
\texttt{DiagnosticProbe} was developed and validated on corpus-grounded QA
tasks with the Qwen3 generator family, so transfer to other task types or
generator families requires separate empirical validation.




\section{Conclusion}
\label{sec:conclusion}

CuratorKIT is a unified and configurable framework for LLM post-training
data curation that brings together source ingestion, schema detection,
data hygiene, synthetic generation, quality filtering, adaptive recovery,
and training-ready export within a single auditable pipeline. Every
sample carries an append-only provenance chain, and every rejected sample
retains a structured record of why it failed rather than being silently
discarded. The companion study~\cite{curatorkit_emnlp26} validates the
core gating and recovery components, finding that exact source provenance
is necessary for reliable hallucination detection, that hallucination and
reward gates capture largely disjoint failure populations making both
necessary, and that adaptive recovery provides its clearest benefit when
smaller generators are non-negotiable. CuratorKIT is compatible with
AlignTune~\cite{aligntune2026}, TRL~\cite{trl2024}, and
Unsloth~\cite{unsloth}.

CuratorKIT will be open-sourced and made available at \url{https://github.com/Lexsi-Labs/CuratorKIT}.


\bibliographystyle{unsrt}
\bibliography{references}

\end{document}